\documentclass[10pt,twocolumn,letterpaper]{article}
\usepackage{pifont}
\usepackage{cvpr}
\usepackage{times}
\usepackage{epsfig}
\usepackage{graphicx}
\usepackage{amsmath}
\usepackage{amssymb}
\usepackage[ruled,vlined]{algorithm2e}
\usepackage{booktabs}
\usepackage{color}
\usepackage{multirow}
\usepackage{amsmath}
\usepackage{bm}
\usepackage{url}
\usepackage{bbding}
\usepackage{indentfirst}
\usepackage{bbm}
\usepackage[T1]{fontenc}
\usepackage{cite}
\usepackage{epstopdf}
\usepackage[title]{appendix}

\usepackage[breaklinks=true,bookmarks=false]{hyperref}

\cvprfinalcopy 

\ifcvprfinal\fi
\begin{document}

\title{Uncertainty-aware Score Distribution Learning for Action Quality Assessment}

\author{Yansong Tang\textsuperscript{1,2,3,}\thanks{\;indicates equal contribution, the corresponding author is Jiwen Lu.},
    Zanlin Ni\textsuperscript{1,}$^*$,
    Jiahuan Zhou\textsuperscript{5},    
    Danyang Zhang\textsuperscript{1},
	Jiwen Lu\textsuperscript{1,2,3},
	Ying Wu\textsuperscript{5},
	Jie Zhou\textsuperscript{1,2,3,4}\\	
\textsuperscript{1} Department of Automation, Tsinghua University, China\\
\textsuperscript{2} State Key Lab of Intelligent Technologies and Systems, China\\
\textsuperscript{3} Beijing National Research Center for Information Science and Technology, China\\
\textsuperscript{4} Tsinghua Shenzhen International Graduate School, Tsinghua University, China\\
\textsuperscript{5} Electrical and Computer Engineering Department, Northwestern University\\
	{\tt \small \{tys15,nzl17,zhang-dy16\}@mails.tsinghua.edu.cn;jiahuanzhou2013@u.northwestern.edu;}\\
	{\tt \small \{lujiwen,jzhou\}@tsinghua.edu.cn;yingwu@eecs.northwestern.edu}}
\maketitle
\thispagestyle{empty}

\begin{abstract}
Assessing action quality from videos has attracted growing attention in recent years.
Most existing approaches usually tackle this problem based on regression algorithms,
which ignore the intrinsic ambiguity in the score labels caused by multiple judges or their subjective appraisals.
To address this issue,
we propose an uncertainty-aware score distribution learning (USDL) approach for action quality assessment (AQA).
Specifically, we regard an action as an instance associated with a score distribution, 
which describes the probability of different evaluated scores.
Moreover, under the circumstance where fine-grained score labels are available (\textit{e.g.,} difficulty degree of an action or multiple scores from different judges),
we further devise a multi-path uncertainty-aware score distributions learning (MUSDL) method to explore the disentangled components of a score.
We conduct experiments on three AQA datasets containing various Olympic actions and surgical activities, 
where our approaches set new state-of-the-arts under the Spearman's Rank Correlation.\footnote{Code is available at \url{https://github.com/nzl-thu/MUSDL}}
\end{abstract}

\section{Introduction}
Action quality assessment (AQA), aiming to evaluate how well a specific action is performed, has become an emerging and attractive research topic in computer vision community because of its potential value for various real-wide applications such as sport video analysis~\cite{DBLP:conf/eccv/PirsiavashVT14,DBLP:journals/corr/abs-1904-04346,DBLP:conf/iccv/BertasiusPYS17a,DBLP:conf/cvpr/ParmarM17}, health care~\cite{DBLP:journals/pami/ZhangL15,DBLP:conf/miccai/FawazFWIM18,DBLP:journals/cars/ZiaSBSE18} and many others~\cite{DBLP:conf/cvpr/DoughtyMD19,DBLP:conf/cvpr/DoughtyDM18,DBLP:journals/cars/ZiaE18,DBLP:journals/cars/ZiaSBSPCE16}.
%
%
Compared with the conventional action recognition problem~\cite{DBLP:conf/cvpr/CarreiraZ17,DBLP:conf/iccv/TranBFTP15,DBLP:conf/cvpr/WangKSL11,DBLP:conf/nips/SimonyanZ14,DBLP:conf/eccv/WangXW0LTG16,DBLP:journals/ivc/Poppe10,DBLP:journals/corr/abs-1812-03982,DBLP:conf/cvpr/TangTLL018,DBLP:journals/tip/TangLWYZ19} focusing on correctly classifying the action sequences from different categories, AQA is a more challenging task as it requires to deal with the videos from the same category with poor intra-class discriminant.


\begin{figure}
\includegraphics[width = \linewidth]{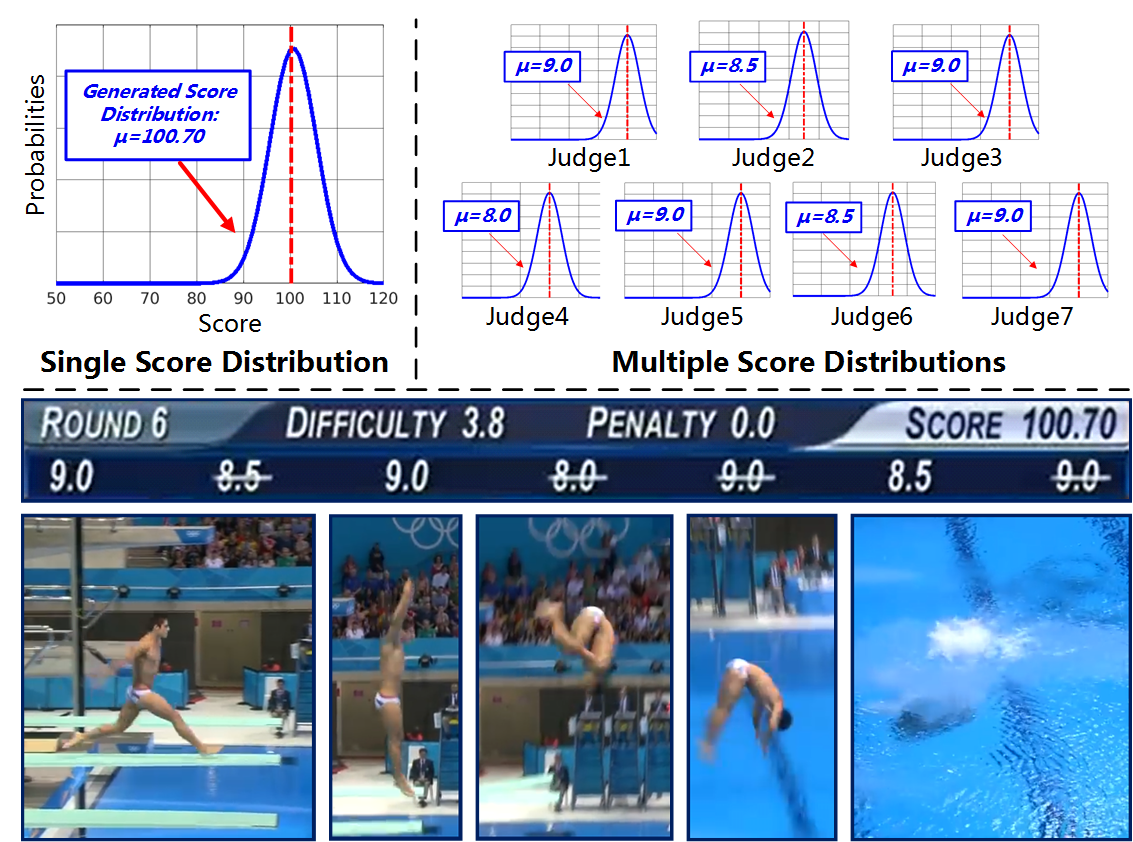} 
\setlength{\belowcaptionskip}{0.05cm}
\caption{We study the problem of action quality assessment in this paper. The bottom row shows a diving action in Olympic game,
in which the final score is calculated based on multiple judges and the difficulty degree as $100.70 = (9.0 + 9.0 + 8.5) \times 3.8$. 
In order to address the uncertainty during the assessment process, 
we utilize a Gaussian distribution to model the final score (top left) and multiple Gaussian distributions to model the scores from different judges (top right).
All figures are best viewed in color.
}
\label{fig:insight}  
\end{figure}

Over the past few years, there have been numbers of methods proposed for AQA~\cite{Gordon-AndrewS1995AIED_AQA,Jug-Marko2003Springer_TrajBasedAQA,Pervse-Matej2007Citeseer_BayesianNet,DBLP:conf/eccv/PirsiavashVT14,DBLP:conf/cvpr/ParmarM17,DBLP:conf/wacv/ParmarM19}. However, most of them simply regard AQA as a regression problem in order to directly predict the action scores.
And unfortunately, their performances are indeed limited.
The root of such a limitation is that this kind of treatment disregards for the underlying ambiguity of the action score labels, which is one of the key issues for AQA. Such ambiguity is caused by how the action labels are generated in practice. As an example shown in Figure~\ref{fig:insight}, for the diving game, when the athlete finishes his action with the difficulty degree of 3.8, seven judges give their scores as $\{9.0, 8.5, 9.0, 8.0, 9.0,  8.5, 9.0\}$.
After eliminating the top-two and bottom-two scores, the final score is calculated as:
$
s_{final} = (9.0+9.0+8.5) \times 3.8 = 100.70. 
$
This suggests the inherent uncertainty of the final score caused by different judges.
Moreover, the subjective appraisal of each individual judge might also bring uncertainty into the final score.
%
%
Besides the diving game, these phenomena exist in many other sports like gymnastic vaults, figure skiing, \textit{etc}. The complicated score uncertainty makes an accurate AQA pretty difficult. Hence, it is desirable to design a robust model to deal with the uncertainty for AQA.

To address this, we propose an uncertainty-aware score distribution learning (USDL) method which utilizes a distribution of different scores as the supervisory signal rather than a single score.
%
The adopted score distribution can depict the probability of the AQA score better so that the aforementioned uncertainty issue can be well handled. 
As illustrated by the top-left of Figure~\ref{fig:insight}, we generate the ground-truth score distribution based on the widely used Gaussian function, of which the mean is set to be the score label.
Meanwhile, an action video is fed into a 3D ConvNets to produce its predicted score distribution.
Then we optimize the Kullback-Leibler divergence between the ground-truth score distribution and the predicted one.
Recall Figure~\ref{fig:insight}, once the fine-grained score labels are available (\textit{e.g.,} difficulty degree of an action or multiple scores from different judges), we further design a multi-path uncertainty-aware score distributions learning (MUSDL) method to fully explore these disentangled components of the final score.
Strictly abided by the rule of the game, we fuse the multiple predicted scores to obtain the final score during inference.
Through this objective process, we are able to obtain more accurate results.
%
To our best knowledge, this is the original effort to leverage the finer-level score annotation for AQA problem.
In order to verify the effectiveness of our approach, we conduct extensive experiments on the AQA-7~\cite{DBLP:conf/wacv/ParmarM19}, MTL-AQA~\cite{DBLP:journals/corr/abs-1904-04346} and JIGSAWS~\cite{gao2014jhu} datasets. Experimental results show the superiority of our methods compared with the state-of-the-arts, and demonstrate the advantage of utilizing fine-grained labels.


\section{Related Work}
\noindent \textbf{Action Quality Assessment:}
In the past years, there have been a variety of works dedicated to different AQA tasks, such as health care~\cite{DBLP:journals/pami/ZhangL15}, instructional video analysis~\cite{DBLP:conf/cvpr/DoughtyMD19,DBLP:conf/cvpr/DoughtyDM18}, sport video analysis~\cite{DBLP:conf/eccv/PirsiavashVT14,DBLP:journals/corr/abs-1904-04346,DBLP:conf/iccv/BertasiusPYS17a,DBLP:conf/cvpr/ParmarM17} and many others\cite{Gordon-AndrewS1995AIED_AQA,Jug-Marko2003Springer_TrajBasedAQA}.
For example,
Pirsiavash \textit{et al.}\cite{DBLP:conf/eccv/PirsiavashVT14} explored this task firstly based on hand-crafted features on several specific actions, they took the first step to apply the learning approach on the underlying task and trained a linear SVR model to regress the score of the videos.
Parmar \textit{et al.} \cite{DBLP:conf/cvpr/ParmarM17} proposed C3D-SVR and C3D-LSTM to predict the score of the Olympic events. Additionally, incremental-label training method was introduced to train the LSTM model based on the hypothesis that the final score is an aggregation of the sequential sub-action scores.
\cite{DBLP:conf/wacv/ParmarM19} and \cite{DBLP:journals/corr/abs-1904-04346} aimed to improve the scoring performance and the generalization ability of the model simultaneously by exploring all-action models and multi-task learning respectively. Meanwhile, both of them released a new AQA dataset with a larger scale.
Xu \textit{et al.}~\cite{score_figure_skating} designed two new LSTM-based models to learn the multi-scale information of the video.
A little different from the works focusing on extracting the whole-scene features, Pan \textit{et al.}~\cite{DBLP:conf/iccv/JiaHuiaction} presented a graph-based model to sufficiently exploit the athletes' pose information. Their approach well balanced the role of the movement of body parts and the coordination among different joints.
Different from the aforementioned regression-based AQA methods, our proposed USDL method aims to predict the score distribution for the input action video instead of a single score number so that the severe score uncertainty which largely limits AQA performance can be well handled.

\begin{figure*}[!h]
\includegraphics[width = \linewidth]{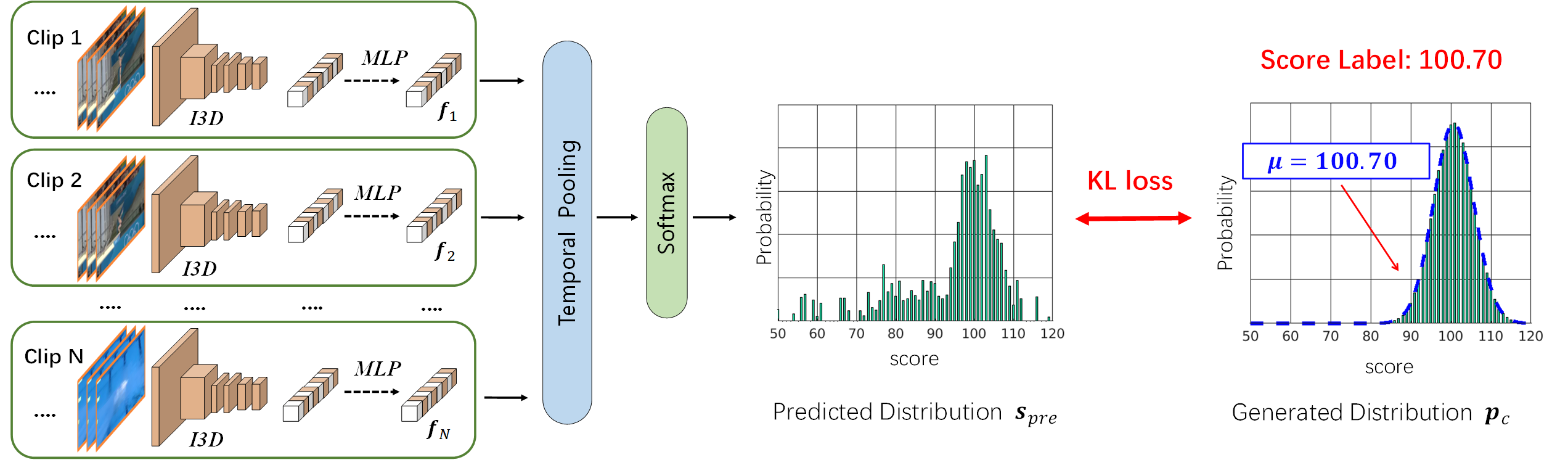} 
\caption{Pipeline of our proposed method for uncertainty-aware score distribution learning.
The input video frames are divided into $N$ segments and fed into an I3D backbone to extract features.
After passing three fully-connected layers, the obtained features are fused by temporal pooling and passed through softmax layer to generate the predicted distribution.
Then we optimize the KL-Loss between the predicted distribution and a Gaussian distribution generated from the score label.}
\label{fig:single-path}  
\end{figure*}

\noindent \textbf{Label Distribution Learning:}
Label distribution learning (LDL) is a general learning paradigm, which describes an instance with distribution rather than the original single label or multiple labels.
As a pioneering work, Geng \textit{et al.} ~\cite{geng2013facial} proposed an LDL framework for facial age estimation, which assigned an age distribution to each face image, and learned from such distribution with two algorithms called IIS-LLD (Improved Iterative Scaling-Learning from Label Distributions) and CPNN (Conditional Probability Neural Network).
Motivated by the success of~\cite{geng2013facial} which was based on the hand-crafted features, the following ameliorated works have been proposed for LDL by leveraging the power of deep learning models such as deep convolutional neural network~\cite{DBLP:journals/tip/GaoXXWG17} or LDL forest~\cite{shen2017label}. 
In recent years, LDL has also shown its effectiveness for various computer vision tasks, including head pose estimation~\cite{DBLP:conf/cvpr/GengX14}, beauty sensing~\cite{DBLP:conf/ijcai/RenG17}, facial landmark detection~\cite{DBLP:conf/aaai/SuG19} and many others~\cite{DBLP:conf/mm/ZhouXG15,DBLP:conf/cvpr/XingGX16,DBLP:conf/ijcai/YangGZ16}.
For video analysis, Geng \textit{et al.} ~\cite{DBLP:conf/aaai/GengL17}  proposed a soft grammar parsing method for video parsing, where the video segments are described by the degrees of different sub-action classes.
Ling \textit{et al.} ~\cite{indoor_counting} utilized a mixture of Gaussian distribution to model the gradual change of crowd numbers in different video frames for indoor crowd counting.
For our proposed method, by transferring the given single score label to a Gaussian-like score distribution for learning, we are able to directly estimate the score distribution of an action video that provides more accurate AQA results than only predicting a single score.

\noindent \textbf{Multi-Label Learning}: Multi-label learning~\cite{tsoumakas2007multi} is increasingly required by more and more computer vision applications including AQA. Recently, A novel MTL-AQA dataset containing fine-grained action labels was proposed by \cite{DBLP:journals/corr/abs-1904-04346}. Besides the final score label of each sequence, the action class and commentary labels are also provided for the usage of multi-label learning. In this work, instead of using extra label information from other tasks, the individual score labels from all judges and the action difficulty label are utilized by our method for multi-label learning with the expectation of a better AQA performance.

\section{Approach}

\subsection{USDL}
\noindent \textbf{Pipeline Overview:}
For a given input video with $L$ frames $V=\{F_l\}_{l=1}^L$, we utilize a sliding window to segment it into $N$ overlapped clips where each clip contains $M$ consecutive frames.
The collected clips are further sent into a backbone of Inflated 3D ConvNets (I3D)~\cite{DBLP:conf/cvpr/CarreiraZ17} followed by three fully connected layers, resulting in $N$ features as $\{\textbf{\textit{f}}_1, \textbf{\textit{f}}_2, ... \textbf{\textit{f}}_N\}$. The weights of the fully connected layers are shared among different clips. As shown in Figure \ref{fig:single-path}, to deal with the extracted features, a straightforward way adopted by the most existing AQA approaches is to fuse them by average pooling or max pooling, and regress it to a final score prediction.
In this work, different from those methods, we utilize a USDL framework to deal with the intrinsic ambiguity in the AQA scores, which will be detailed as below.

\noindent \textbf{Score Distribution Generation:}
During the training phase, given a video associated with the labeled score $s$, we first generate a Gaussian function with the mean of $s$ and standard deviation of $\sigma$ as follow:
\begin{eqnarray}
\label{eqn:gaussian}
g(c) = \frac{1}{\sqrt{2\pi}\sigma}\exp(-\frac{(c-s)^2}{2\sigma^2}). 
\end{eqnarray}

Here $\sigma$ is a hyper-parameter which serves as the level of uncertainty for assessing an action.
By uniformly discretizing the score interval into a set of scores $\textbf{\textit{c}} = [c_1, c_2, ..., c_m]$, a vector is utilized to describe the degree of each score as $\textbf{\textit{g}}_c = [g(c_1), g(c_2), ..., g(c_m)]$. 
The final score distribution label $\textbf{\textit{p}}_c = [p(c_1), p(c_2), ..., p(c_m)]$ is generated by normalizing $\textbf{\textit{g}}_c$ as below:
\begin{eqnarray} 
p(c_i) = g(c_i)/\sum_{j=1}^m g(c_j), \; i = 1, 2, ..., m. 
\end{eqnarray}

\noindent \textbf{Learning from Score Distribution:}
In order to learn from the obtained distribution $\textbf{\textit{p}}_c$, we map the $N$ learned features $\{\textbf{\textit{f}}_1, \textbf{\textit{f}}_2, ..., \textbf{\textit{f}}_N\}$ into $N$ predicted scores as $\{\textbf{\textit{s}}_1, \textbf{\textit{s}}_2, ..., \textbf{\textit{s}}_N\}$, where $\textbf{\textit{s}}_n$ has the same size with the $\textbf{\textit{p}}_c$.
Then, the temporal average pooling is performed to $\{\textbf{\textit{s}}_n\}_{n=1}^N$ for an output vector $\textbf{s}'$.
After the softmax activation on $\textbf{s}'$, we obtain the final predicted score as $\textbf{s}_{pre} = [s_{pre}(c_1), s_{pre}(c_2), ..., s_{pre}(c_m)]$. 
Finally, the learning loss is calculated as the Kullback-Leibler (KL) divergence between $\textbf{s}_{pre}$ and $\textbf{\textit{p}}_c$:
\begin{eqnarray} 
KL\{\textbf{\textit{p}}_c|| \textbf{s}_{pre}\} = \sum_{i=1}^m p(c_i) \log \frac{p(c_i)}{s_{pre}(c_i)}.
\end{eqnarray}

\noindent \textbf{Inferring from Score Distribution:}
During the inferring stage, we forward the input testing video into our optimized model to obtain the corresponding predicted score distribution $\textbf{s}_{pre}$. The final assessment is obtained by selecting the score with the max probability:
\begin{eqnarray} 
s_{final} = \mathop{\arg\max}_{c_i} \ \  \{s_{pre}(c_1), s_{pre}(c_2), ..., s_{pre}(c_m)\}. 
\end{eqnarray}

\begin{figure}
\includegraphics[width = \linewidth]{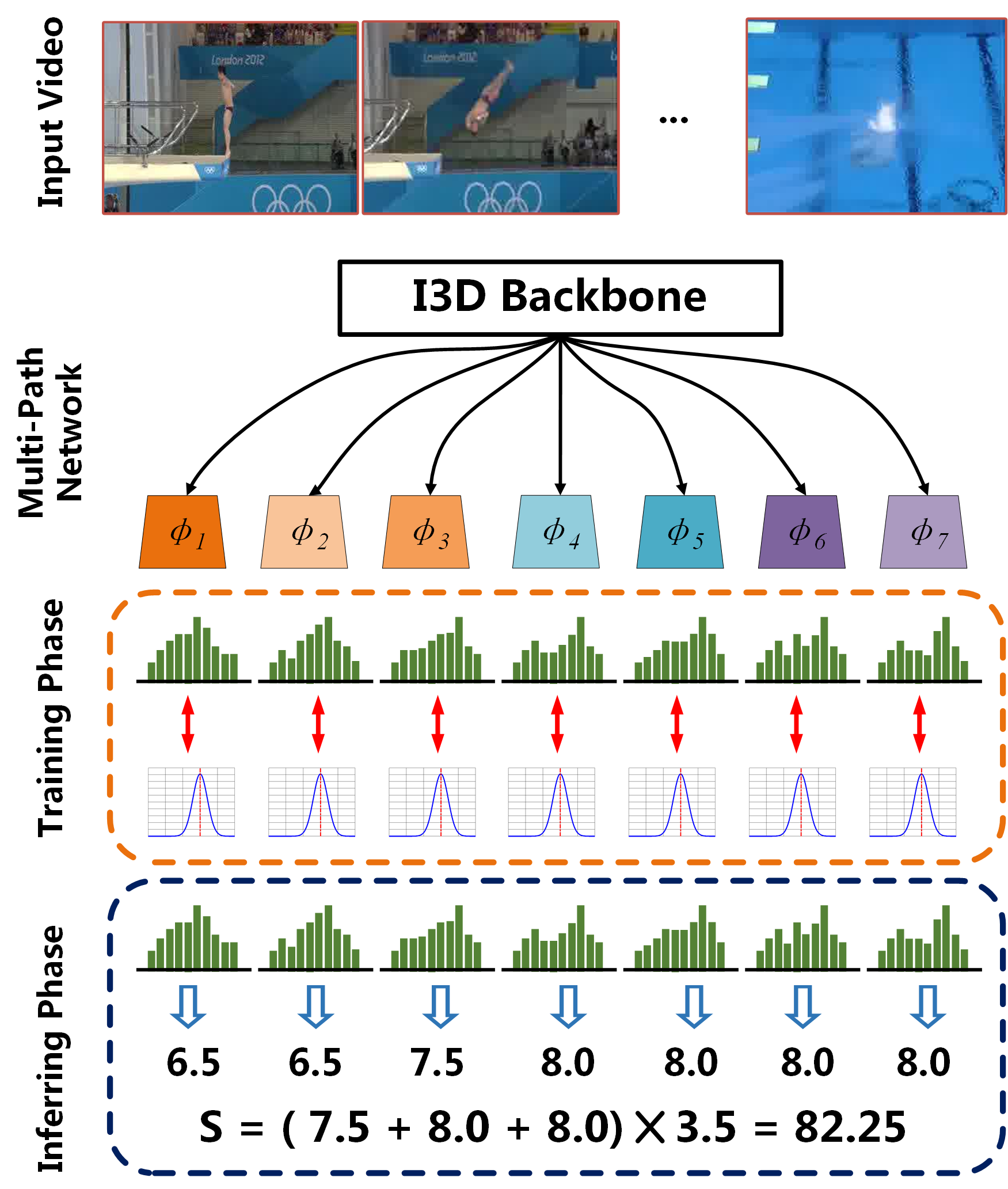}
\setlength{\belowcaptionskip}{0.05cm}
\caption{Multi-path uncertainty-aware score distributions learning. During the training phase, we model the scores from $K$ judges as different Gaussian distributions and utilize a similar strategy to train the model containing $K$ sub-networks. During the testing phase, we obtain the final assessment based on the $K$ predicted scores and the rule of the game.}
\label{fig:multi-path}  
\vspace{-0.2cm}
\end{figure}

\subsection{MUSDL}
In most existing works on AQA, the network is designed and optimized based on the single score supervision.
However, in many Olympic games (\textit{e.g.,} diving, figure skiing, \textit{etc.}), the final score is calculated based on multiple scores from several judges according to a certain rule.
Thanks to the recently released MTL-AQA dataset, the raw annotation of different judges and the difficulty degree are available as the intermediate components.
In this subsection, we further introduce a multi-path architecture to leverage these intermediate components for AQA.

\noindent \textbf{Learning from Multi-path Score Distributions:}
As shown in Figure \ref{fig:multi-path}, for each single-path, we use the same pipeline as our USDL approach. The fully connected layers of different paths are separately trained while the I3D backbone is shared among paths.
In the training phase, suppose we have a set of scores $\{s^{judge}_k\}_{k=1}^K$ from $K$ different judges. We first sort the scores in increasing order in order to train the sub-networks representing judges of different rigor.
Following Eqn~(\ref{eqn:gaussian}), we generate $K$ Gaussian distributions as $\{\textbf{\textit{p}}_{c,k}^{judge}\}_{k=1}^K$.
Given a training video, we first feed it through the I3D backbone and obtained the $N$ features as $\{\textbf{\textit{f}}_1, \textbf{\textit{f}}_2, ... \textbf{\textit{f}}_N\}$. 
The features are then fed into $K$ sub-networks to obtain $K$ final predicted distributions $\{\textbf{s}_{pre,k}^{judge}\}_{k=1}^K$ as follow:
\begin{eqnarray} 
\textbf{s}_{pre,k}^{judge} =  \phi_k(\textbf{\textit{f}}_1, \textbf{\textit{f}}_2, ... \textbf{\textit{f}}_N),  \quad k = 1, 2, ... K.
\end{eqnarray}

Then the total training loss is calculated as: 
\begin{eqnarray} 
J_{multi} &=& \sum_{k=1}^K KL\{\textbf{\textit{p}}_{c,k}^{judge} ||\textbf{s}_{pre,k}^{judge}\} \\ \nonumber
                  &=& \sum_{k=1}^K \sum_{i=1}^m p(c_{i, k}^{judge}) \log \frac{p(c_{i, k}^{judge})}{s^{judge}_{pre,k}(c_{i, k}^{judge})}.
\end{eqnarray}

\noindent \textbf{Rule-based Multi-path Inference:}
During the inferring phase, we forward each testing video through our multi-path model and obtain $K$ final predicted scores as $\{s_{final,k}^{judge}\}_{k=1}^K$.
According to the rule specific to the diving game, we can obtain the final score as:
\begin{eqnarray} 
\label{s_final}
s_{final} = DD \times \sum_{k' \in U } s_{final,k'}^{judge}.
\end{eqnarray}

Here $U$ denotes a subset of $\{1, 2, .., K\}$ (\textit{e.g.,} the diving game would discard the judges giving the top 2 and the last 2 scores), and $DD$ denotes the difficulty degree of the input action video which will be released in advance.
In fact, even the $DD$ is not provided during inference time, we can still train a model to predict $DD$ by introducing a side-network branch for it during training. The predicted $DD$ is directly used for Eqn~(\ref{s_final}) during inference.
We will report the results of these two cases ($DD$ is available or unavailable during testing) in the Experiment Section.

\section{Experiment}
\subsection{Datasets and Experiment Settings}
\noindent \textbf{AQA-7}\cite{DBLP:conf/wacv/ParmarM19}: 
The AQA-7 dataset contains totally 1189 samples from 7 sports: 
370 from \textit{single diving - 10m platform}, 176 from \textit{gymnastic vault}, 175 from \textit{big air skiing}, 206 from \textit{big air snowboarding}, 88 from \textit{synchronous diving - 3m springboard}, 91 from \textit{synchronous diving - 10m platform} and 83 from \textit{trampoline}.
We followed the setting in \cite{DBLP:conf/wacv/ParmarM19} and excluded the trampoline category in which the videos are much longer than those in the other categories~\cite{DBLP:conf/iccv/JiaHuiaction}. 
There were 803 clips for training and 303 clips for testing.

\noindent  \textbf{MTL-AQA}\cite{DBLP:journals/corr/abs-1904-04346}:
The MTL-AQA dataset is the currently largest dataset for AQA.
There are 1412 fine-grained samples collected from 16 different events with various views in MTL-AQA. This dataset covers the events of both individual \& synchronous divers, both male \& female athletes, both 3m springboard \& 10m platform settings. 
In this dataset, different kinds of annotations are provided to enable the study for different tasks consisting of action quality assessment, action recognition and commentary generation.
Furthermore, the raw annotations of the scores from seven judges and the difficulty degree ($DD$) are available for each action.
We followed the evaluation protocol suggested in \cite{DBLP:journals/corr/abs-1904-04346} to divide the dataset into a 1059-sized training set and a 353-sized test set.

\noindent \textbf{JIGSAWS}
\cite{gao2014jhu}:
%
Besides sport events, 
we further evaluate our methods on the JIGSAWS~\cite{gao2014jhu} dataset which contains surgical activities.
There are 3 tasks as ``Suturing (S)'', ''Needle Passing (NP)'' and ``Knot Tying (KT)'' in this dataset.
Since each video is annotated with multiple annotation scores assessing different aspects of a video (\textit{e.g.,} flow of operation, quality of final product, \textit{etc.}) and the final score is defined as the sum of these sub-scores,
we could easily extend our MUSDL based on this rule.
There are stereo samples recorded by left and right cameras in this dataset, and we only used the videos from the left view due to the high similarity between the paired captures.
We adopt a similar four-fold cross validation~\cite{DBLP:conf/iccv/JiaHuiaction,DBLP:conf/cvpr/DoughtyDM18}. Since the concrete splits are not public available for~\cite{DBLP:conf/iccv/JiaHuiaction}, and the splits in \cite{DBLP:conf/cvpr/DoughtyDM18} are specifically designed for pair-wise rank evaluation, which are not suitable for our work and~\cite{DBLP:conf/iccv/JiaHuiaction}, we simply divided the dataset into four folds randomly. Concretely, our splits are released at the project page.

\noindent \textbf{Evaluation Protocols:}
To keep alignment with existing literatures~\cite{DBLP:conf/iccv/JiaHuiaction}, we used Spearman's rank correlation (ranging from -1 to 1, the higher the better) to measure the performance of our methods between the ground-truth and predicted score series. Spearman's correlation is defined as:
\begin{eqnarray}
\rho = \frac {\sum_i(p_i-\bar p)(q_i-\bar q)}{\sqrt{\sum_i(p_i-\bar p)^2\sum_i(q_i-\bar q)^2}}.
\end{eqnarray}

Here $p$ and $q$ represent the ranking of two series respectively. Fisher's z-value~\cite{DBLP:conf/wacv/ParmarM19} is used to measure the average performance across actions.
\subsection{Implementation Details}
Our proposed methods were built on the Pytorch toolbox~\cite{paszke2017automatic} and implemented on a system with the Intel(R) Xeon(R) E5-2660 v4 CPU @ 2.00Ghz.
We trained our model with two Nvidia GTX 1080 Ti GPUs.
We utilized the I3D model pretrained on Kinetics dataset as a feature extractor.
It took action sequences containing 16 frames as inputs and output a feature with 1024 dimensions. In AQA-7 and MTL-AQA, the videos are all 103 frames long. We divided each video into ten segments under a certain rule\footnote{We took No. [0, 10, 20, 30, 40, 50, 60, 70, 80, 87] frame as the beginning frame for ten segments. See supplementary material for details.}. The MLP block, containing two hidden layers $FC(256,ReLU)$ and $FC(128,ReLU)$, together with temporal pooling layer and softmax layer, built the score distribution for each video.
We performed temporal pooling at feature-level on MTL-AQA and JIGSAWS for better results.
Adam~\cite{Kingma2014Adam} was adopted for network optimization.
In our experiments, we normalized the final total score in both datasets and seven judge scores in MTL-AQA. For the final total score, since it was a float number, we normalized it as:
\begin{eqnarray} 
S_{normalize} = \frac{S - S_{min}}{S_{max} - S_{min}} \times 100.
\end{eqnarray}

Here $S_{max}$ and $S_{min}$ indicate the maximum and minimum score in the dataset.
For judge scores in MTL-AQA dataset, since these scores are inherently discrete but not an integer, so we normalized them by doubling the value of original score to get an integer.
After generating normalized score $S_{normalized}$, we produced a Gaussian function with a mean of $S_{normalized}$. 
Note that the produced Gaussian is defined within $x \in (-\infty, \infty)$ in first. In practice, we truncated this initial distribution by the score range, and then discretized and renormalized this distribution function as described in Section 3.1. 

%
We report the performance of the following baseline methods and different versions of our approach. Note that some of them are not evaluated on the AQA-7 dataset due to the absense of multiple scores.
 
\begin{itemize}
\item Regression: 
Most existing works adopted this strategy. 
We modified the dimension of the last fc layer in our USDL to produce a single prediction score.
During the training phase, we optimized the L2 loss between predicted score and ground-truth score.

\item USDL:
The proposed method in Section 3.1.

\item MUSDL and MUSDL$^*$:
The proposed methods in Section 3.2, which used the ground-truth and the prediction of difficulty degree during testing respectively.

\item USDL$_{DD}$:
During the training phase,
we used scores from seven judges. 
According to the scoring rules of diving, the top-two and bottom-two scores will be eliminated. 
We summed the remaining three judge scores to obtain a new score label,
and applied USDL to learn this new label.
In the inference period, we multiply the predicted score with the ground-truth of difficulty degree $DD$ to generate the final result.
\end{itemize}

\begin{figure*}[th]
\centering
\includegraphics[width = 0.95\linewidth]{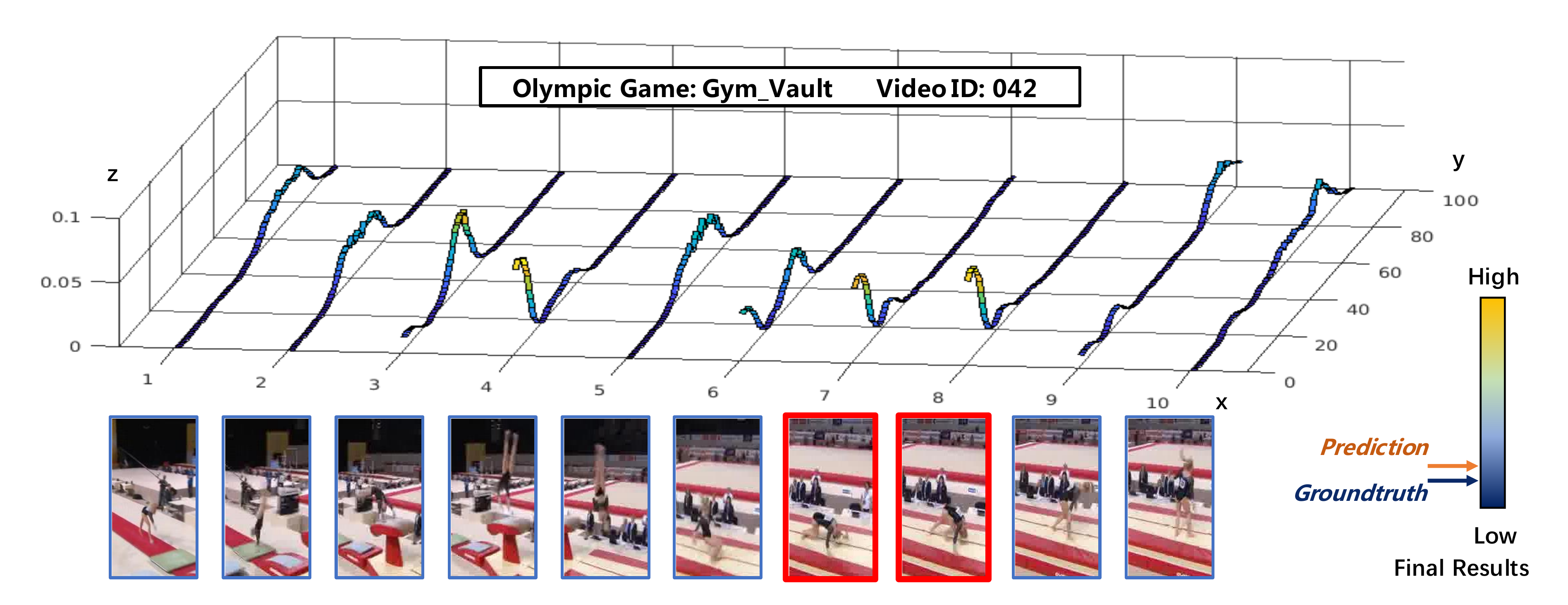} 
\caption{Score distribution change in temporal domain. The x, y, z axis represents clip number, score and probability predicted for the certain score. The 7th, 8th clips are highlighted because the athlete fell to the ground, which leads to two salient low score prediction.
}
\label{fig:tem_score}  
\end{figure*}

\begin{table*}[!t]
\small
\caption{Comparisons of action quality assessment accuracy on the AQA-7 dataset.} \label{tab:aqa_cm}
\vskip 0.1 in
\linespread{1.4}
\centering
\begin{tabular}{l | c c c c c c | c }
\toprule[1.5pt]
 & Diving & Gym Vault & Skiing & Snowboard & Sync. 3m & Sync. 10m & Avg. Corr.\\
\midrule[1.2pt] 
Pose+DCT~\cite{DBLP:conf/eccv/PirsiavashVT14} & 0.5300 & -- & -- & -- & -- & -- & -- \\
ST-GCN~\cite{DBLP:conf/aaai/YanXL18} & 0.3286 & 0.5770 & 0.1681 & 0.1234 & 0.6600 & 0.6483 & 0.4433 \\
C3D-LSTM~\cite{DBLP:conf/cvpr/ParmarM17} & 0.6047 & 0.5636 & 0.4593 & 0.5029 & 0.7912 & 0.6927 & 0.6165 \\
C3D-SVR~\cite{DBLP:conf/cvpr/ParmarM17} & 0.7902 & 0.6824 & 0.5209 & 0.4006 & 0.5937 & 0.9120 & 0.6937 \\
JRG~\cite{DBLP:conf/iccv/JiaHuiaction} & 0.7630 & 0.7358 & 0.6006 & 0.5405 & 0.9013 & \textbf{0.9254} & 0.7849 \\
\hline
Ours-Regression & 0.7438 & 0.7342  & 0.5190 & 0.5103 & 0.8915 & 0.8703 & 0.7472 \\
Ours-USDL & \textbf{0.8099} & \textbf{0.7570} & \textbf{0.6538} & \textbf{0.7109} & \textbf{0.9166} & 0.8878 & \textbf{0.8102} \\
\bottomrule[1.5pt]
\end{tabular}
\end{table*}
\linespread{1}

\begin{table*}[!t]
\small
\caption{Study on different distributions. We use USDL approach with only the type of soft distribution changed.} \label{tab:dis_study}
\vskip 0.1 in
\linespread{1.3}
\centering
\begin{tabular}{l | c c c c c c | c }
\toprule[1.5pt]
 & Diving & Gym Vault & Skiing & Snowboard & Sync. 3m & Sync. 10m & Avg. Corr.\\
\midrule[1.2pt] 
$\chi^2$ Distribution & 0.7920 & \textbf{0.7697} & 0.6532 & 0.6905 & 0.9041 & 0.8847 & 0.8015 \\
Triangle Distribution & \textbf{0.8147} & 0.7452 & 0.5960 & 0.6555 & 0.8987 & \textbf{0.9054} & 0.7969 \\
Gaussian Distribution & 0.8099 & 0.7570 & \textbf{0.6538} & \textbf{0.7109} & \textbf{0.9166} & 0.8878 & \textbf{0.8102} \\
\bottomrule[1.5pt]
\end{tabular}
\end{table*}

\subsection{Results on AQA-7 Dataset}
Table \ref{tab:aqa_cm} shows the experiment results of our methods and the comparison with other AQA approaches.
The simple yet effective regression model achieves competitive performance compared with the state-of-the-art C3D-LSTM, C3D-SVR and JRG models. The proposed USDL approach obtains significant improvement for all action classes except \textit{Sync. 10m} compared to the JRG model, which also utilizes optical flow. The average correlation of our USDL approach gains improvement of 6.3\% and 2.5\% compared to our baseline regression model and the JRG model respectively. It is also noticed that in action class \textit{Snowboard}, our proposed model surpasses state-of-the-art models by a large margin about 17\%, showing the strong effectiveness of the USDL approach in AQA problems.


\noindent \textbf{Study on Different Distributions:}
We conduct 3 parallel experiments with only the type of distribution changed. The $\chi^2$ distribution is generated where the mean $s$ equals to normalized score and degree of freedom equals to one-twentieth of score range. The triangle distribution is symmetric with maximum probability at normalized score. The Gaussian distribution follows the implementation in Section 4.2. And all of the distributions are truncated discretized and renormalized to fit the range of scores. As shown in Table \ref{tab:dis_study}, Gaussian distribution achieves the highest average correlation while the triangle distribution performs worst. However, note that there is not a distribution that performs best in all action classes. For example, triangle distribution performs best in \textit{Diving} class and $\chi^2$ distribution performs best in \textit{Gym Vault} class, while the Gaussian distribution performs best in an average sense. This indicates that the ground-truth score distribution may be complicated and vary a lot among different action classes.

\noindent \textbf{Visualization of Temporal Evolution:}
We choose a video in \textit{Gym Vault} class to visualize the temporal evolution of score distribution in Figure~\ref{fig:tem_score}. We obtain the score distributions of 10 clips and plot them on the graph. The video has a low ground-truth score and network gives a low score prediction. From the temporal evolution of score distribution we can see how the network gives that prediction---in the 7th and the 8th segments, the athlete fell to the ground, leading to two salient low-score prediction. The two distributions thus dominate in the final distribution, and the network finally gives the low score distribution, as expected. See supplementary material for more visualization results.

\begin{figure*}[!t]
\centering
\includegraphics[width = 0.95\linewidth]{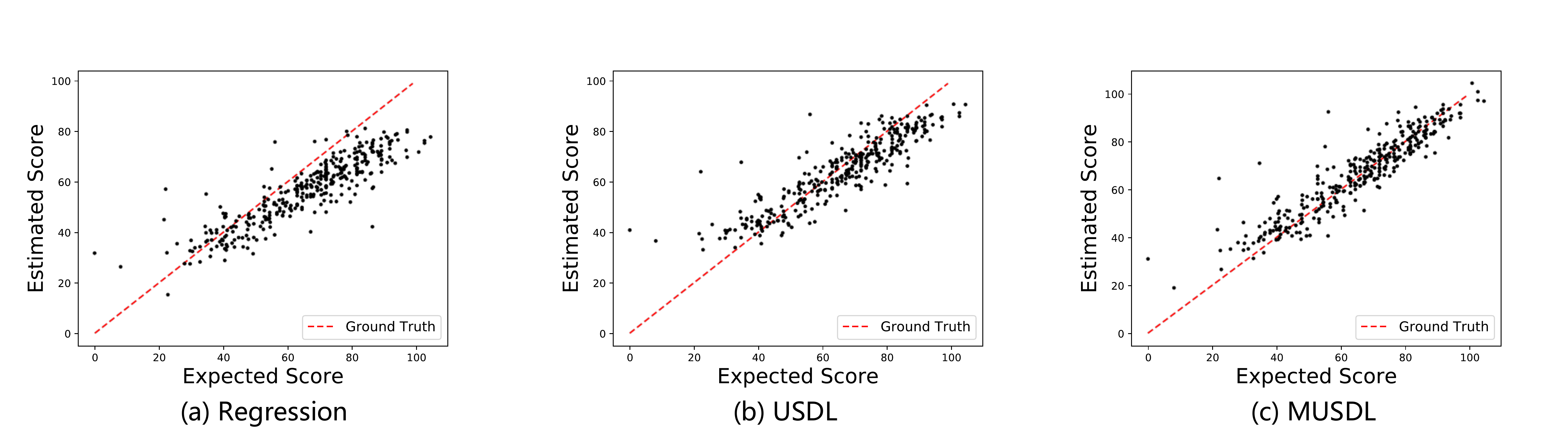} 
\caption{A comparison of different methods in scatter plot. The plotting points are the predictions of network with y-coordinates being the predicted scores and x-coordinates being the ground-truth scores. The ground-truth samples are plotted in dotted line.}
\label{fig:scatter}  
\end{figure*}

\subsection{Results on MTL-AQA Dataset}

\begin{table}[!t]
\small
\caption{Comparisons of performance with existing methods on the MTL-AQA dataset.} \label{tab:mtl_cm}
\vskip 0.1 in
\linespread{1.2}
\centering
\begin{tabular}{l | c}
\toprule[1.5pt]
Method & Sp. Corr.\\
\midrule[1.2pt] 
Pose+DCT~\cite{DBLP:conf/eccv/PirsiavashVT14} & 0.2682 \\
C3D-SVR~\cite{DBLP:conf/cvpr/ParmarM17} & 0.7716 \\
C3D-LSTM~\cite{DBLP:conf/cvpr/ParmarM17} & 0.8489 \\
MSCADC-STL~\cite{DBLP:journals/corr/abs-1904-04346} & 0.8472\\
C3D-AVG-STL~\cite{DBLP:journals/corr/abs-1904-04346} & 0.8960\\
MSCADC-MTL~\cite{DBLP:journals/corr/abs-1904-04346} & 0.8612\\
C3D-AVG-MTL~\cite{DBLP:journals/corr/abs-1904-04346} & 0.9044\\
\hline
Ours-Regression &  \textbf{0.8905} \\
Ours-MUSDL &   \textbf{0.9273} \\
\bottomrule[1.5pt]
\end{tabular}
\vspace{-0.2cm}
\end{table}
\linespread{1}

Our method is compared with several state-of-the-arts on MTL-AQA dataset in Table \ref{tab:mtl_cm}.
Still, our regression model obtains a competitive result compared with the state-of-the-arts and our MUSDL model outperforms all the other approaches listed out. 
These experiment results convincingly illustrate the effectiveness of our method. It is believed that embedding the MUSDL method into a regression model is able to boost the assessment accuracy.


\begin{table}[!t]
\small
\caption{Ablation study on the MTL-AQA dataset.
For MUSDL$^*$, \ding{51}$^*$ denote that $DD$ is only available during the training phase, but unavailable during testing phase.} \label{tab:mtl_ab}
\vskip 0.1 in
\linespread{1.2}
\centering
\begin{tabular}{l | c c c | c}
\toprule[1.5pt]
Method & Soft & DD & Judges & Sp. Corr.\\
\midrule[1.2pt] 
Regression  & & & & 0.8905\\
USDL & \ding{51} & & & 0.9066  \\
USDL$_{DD}$ &  \ding{51}  &  \ding{51}  &  single & 0.9231 \\
MUSDL$^*$ &  \ding{51}  &  \ding{51}$^*$  & multiple & 0.9158 \\
\hline
MUSDL & \ding{51}  &  \ding{51}  &  multiple &  \textbf{0.9273}  \\
\bottomrule[1.5pt]
\end{tabular}
\vspace{-0.3cm}
\end{table}
\linespread{1}

\begin{figure}[!t]
\centering
\includegraphics[width = 0.9\linewidth]{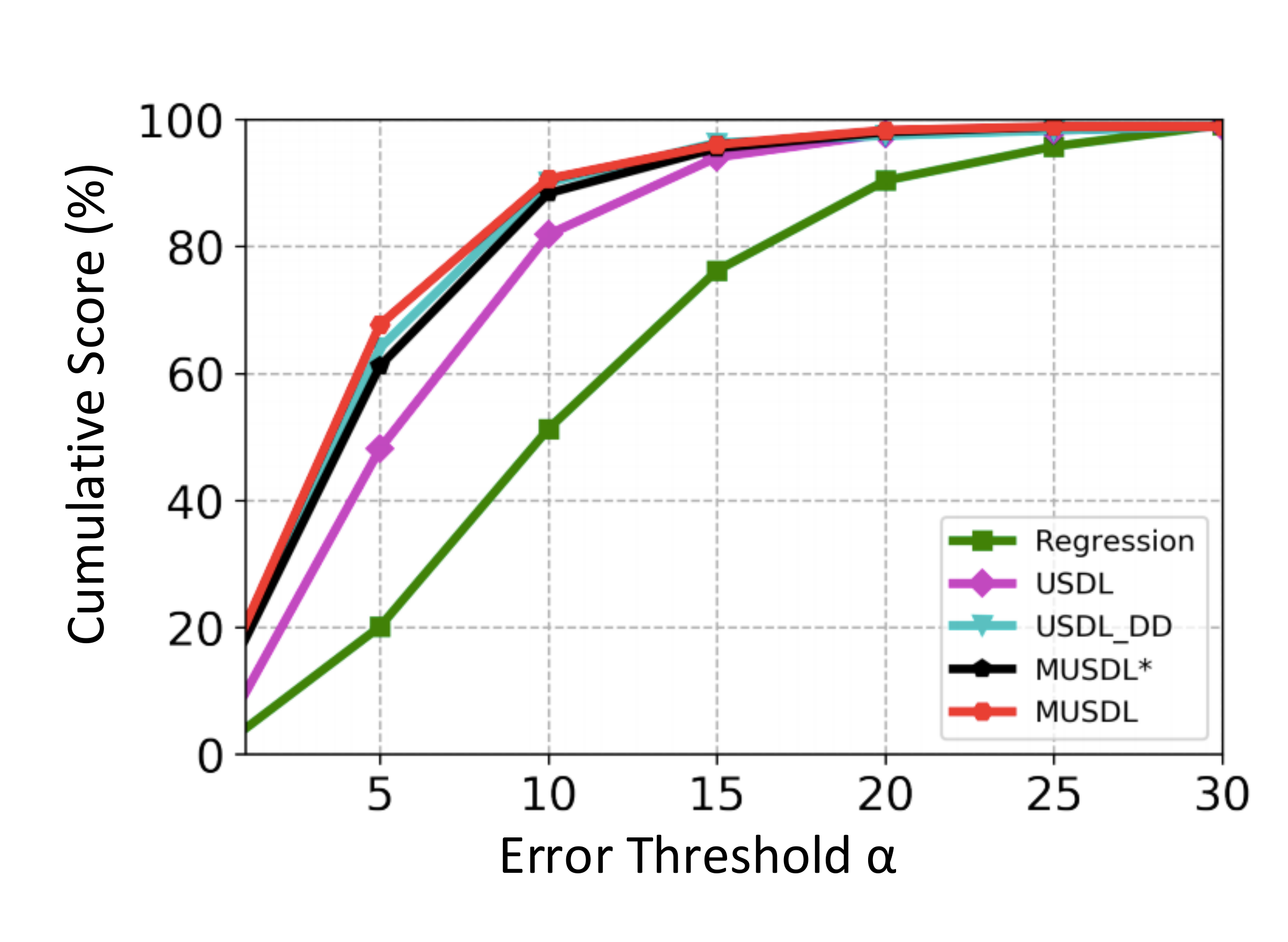}
\caption{Cumulative score curve in MTL-AQA dataset, in which x-axis measures the absolute difference between predition and label and y-axis shows the proportion of the sample within the current error level. The methods using $DD$ outperform baseline methods significantly, and the MUSDL method performs better than other methods using $DD$ by a little margin.}
\label{fig:cs-curve}  
\vspace{-0.3cm}
\end{figure}

\begin{figure*}[!t]
\centering
\includegraphics[width = 0.95\linewidth]{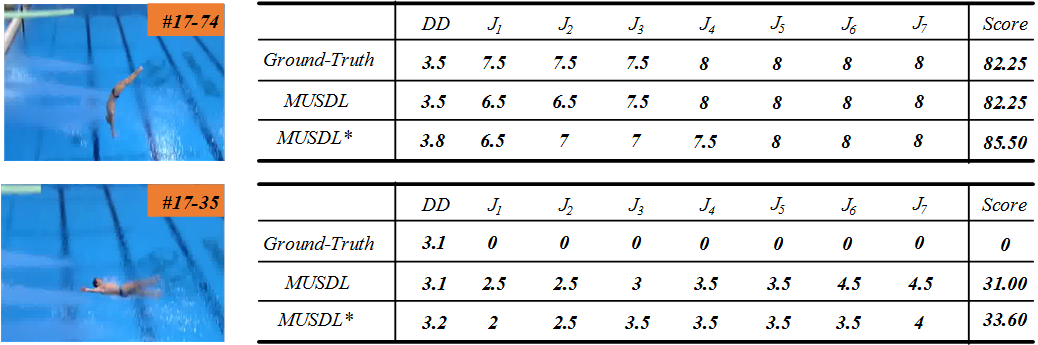}
\setlength{\belowcaptionskip}{0.05cm}
\caption{Case study with qualitative results, which present the comparisons of MUSDL, MUSDL* and ground-truth. 
}
\label{fig:case_study}  
\vspace{-0.1cm}
\end{figure*}

\noindent \textbf{Ablation Study:}
We compare our final method with 4 other methods as shown in Table \ref{tab:mtl_ab}. The vanilla regression and USDL methods do not exploit the information from $DD$ nor the judge scores. 
The architecture of USDL$_{DD}$ network is the same as USDL, but the distribution label is generated from multiple judge scores. The final prediction is obtained by combining $DD$ and predicted judge scores, which is the same as MUSDL method.
The MUSDL* method adds an additional branch to the previous MUSDL method to perform a multi-task learning, \textit{i.e.,} training the network to predict scores from seven judges and $DD$ at the same time. In inference stage, we combine the judge scores and $DD$ predicted from network to obtain the final score.
From the results we see that the USDL$_{DD}$ outperforms the USDL by 1.7\%, indicating that $DD$ is a significant factor in diving score assessment. We believe the reason why using $DD$ can promote better performance is that it ``disentangles'' the problem, making the main pipeline more specialized in video quality assessment. The MUSDL outperforms the single path method by 0.4\%, which shows the fine-grained scores can further improve the performance of network.

\noindent \textbf{Visualization:}
We first evaluate the methods in ablation study using cumulative score (CS) curves as shown in Figure \ref{fig:cs-curve}. The cumulative prediction accuracy at the error $\epsilon$ is computed as
$
CS(\alpha) = \frac {N_{\epsilon \leq \alpha} } N\times 100\%.
$
Here $N_{\epsilon\leq \alpha}$ is the number of videos on which the prediction error $\epsilon$ is not larger than the threshold $\alpha$.
The figure shows the strong effectiveness of using $DD$ for learning better distributions. 

Then we choose the Regression, USDL and MUSDL methods and plot the scatter diagram to make a further comparison, as shown in Figure~\ref{fig:scatter}. The predicted scores are plotted in scatter points and the ground-truth scores are plotted in dotted line. From the results, we can see that the proposed MUSDL obtains a very satisfactory performance since the predicted points converge well to the ground-truth line.

We also conduct a case study to analyze the behavior of our network, as shown in Figure~\ref{fig:case_study}. The first case is a successful case where the final score predicted by MUSDL fully matches with ground-truth, and final score predicted by MUSDL* network is a little higher than ground-truth. The second case is a failure case, with the ground-truth label of the video being zero. This time both of our MUSDL network and MUSDL* network give the score far higher than the ground-truth label. The results indicate that when assessing videos with extreme scores, \textit{i.e.,} too far from the normal score level, the proposed method may not be able to get satisfactory results. This is because the samples with extreme scores are scarce and also because the intrinsic property of our method makes it harder to give extreme results. 

\begin{table}[t]
\small
\caption{Comparisons of action quality assessment accuracy on the JIGSAWS dataset.} \label{tab:jigsaws}
\vskip 0.1 in
\centering
\begin{tabular}{l | c c c c}
\toprule[1.5pt]
Method & S & NP & KT & Avg. Corr.\\
\midrule[1.2pt] 
ST-GCN~\cite{DBLP:conf/aaai/YanXL18,DBLP:conf/iccv/JiaHuiaction} & 0.31 &0.39& 0.58& 0.43\\
TSN~\cite{DBLP:conf/eccv/WangXW0LTG16,DBLP:conf/iccv/JiaHuiaction} & 0.34 & 0.23 & 0.72 & 0.46\\
JRG~\cite{DBLP:conf/iccv/JiaHuiaction} & 0.36 & 0.54 & \textbf{0.75} & 0.57\\
\hline
Ours-USDL & 0.64 & 0.63 & 0.61 & 0.63\\
Ours-MUSDL & \textbf{0.71} & \textbf{0.69} & 0.71 & \textbf{0.70}\\
\bottomrule[1.5pt]
\end{tabular}
\vspace{-0.1cm}
\end{table}

\subsection{Results on the JIGSAWS dataset}
We finally conduct experiments on the JIGSAWS dataset for surgical activities.
Because the lengths of each videos are much longer that those in AQA-7 and MTL-AQA datasets,
we uniformly sampled 160 frames of each videos and divided them into 10 segments as the inputs of our models.
Table \ref{tab:jigsaws} presents the results of our methods compared with the state-of-the-arts.
Our MUSDL achieves best performance of 0.71 (S), 0.69 (NP), 0.70 (Avg. Corr.) and comparable result of 0.71 (KT), which demostrates its effectiveness for action quality assessment.

\section{Conclusion}
In this paper, we have proposed a new uncertainty-aware score distribution learning (USDL) method for action quality assessment, 
which aims to address the inherent ambiguity in the score label.
Moreover, we have devised a multi-path uncertainty-aware score distribution (MUSDL) framework to take advantage of additional fine-grained score labels. 
The experiments on three AQA datasets have demonstrated the effectiveness of our approach.
In the future,
we plan to apply our methods for instructional video analysis~\cite{coin,coin_pami}.
Besides, it is an interesting direction to explore the interpretability of AQA models (\textit{e.g.,} understanding how the network gains the score for a certain action),
which is virtually important for real-world applications.

\section{Acknowledgement}
This work was supported in part by the National Key Research and Development Program of China under Grant 2017YFA0700802, in part by the National Natural Science Foundation of China under Grant 61822603, Grant U1813218, Grant U1713214, and Grant 61672306, in part by the Shenzhen Fundamental Research Fund (Subject Arrangement) under Grant JCYJ20170412170602564, in part by Tsinghua University Initiative Scientific Research Program, in part by National Science Foundation grant IIS-1619078, IIS-1815561, and in part by the Army
Research Office ARO W911NF-16-1-0138.
The authors would sincerely thank Xumin Yu, Wanhua Li, Jia-Hui Pan and Paritosh Parmar for their generous helps.

\begin{appendices}

\begin{figure*}[!h]
\centering
\includegraphics[width = 0.75\linewidth]{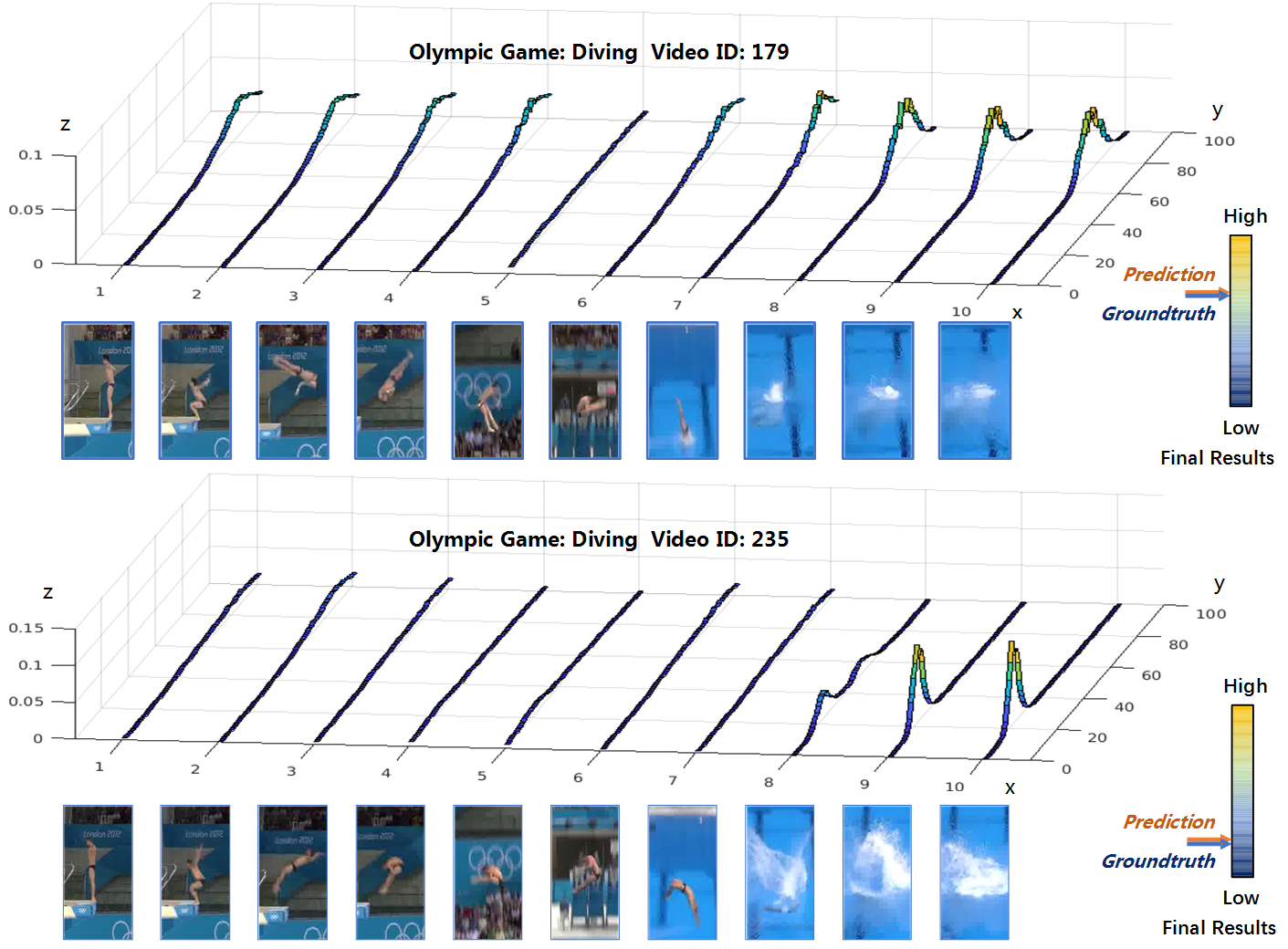} 
\caption{Evolution of the score distributions in temporal domain. 
The x, y, z axis represents clip number, score and probability predicted for the certain score.
The two samples are selected from the \textit{Diving} action in AQA-7~\cite{DBLP:conf/wacv/ParmarM19}.
}
\label{fig:temporal}  
\end{figure*}

\begin{table}[ht]
\small
\caption{Comparisons of different strategies to segment on the \textit{Diving} action of AQA-7~\cite{DBLP:conf/wacv/ParmarM19}.} \label{tab:segment}
\vskip 0.1 in
\centering
\begin{tabular}{l | c c c}
\toprule[1.5pt]
Method & 6-seg~\cite{DBLP:journals/corr/abs-1904-04346} & 10-seg-s1 & 10-seg-s2\\
\midrule[1.2pt] 
Sp. Corr. & 0.7642 & \textbf{0.8099} &  0.7928 \\
\bottomrule[1.5pt]
\end{tabular}
\vspace{-0.3cm}
\end{table}

\section{Different Strategies for Segmenting Videos}
During our experiments, we explored three strategies to divide the videos into multiple segments.
Table \ref{tab:segment} presents the results on the \textit{Diving} action of the AQA-7 dataset~\cite{DBLP:conf/wacv/ParmarM19},
where the length of each video clip is 103 frames.
Specifically, ``6-seg'' denotes the scheme used in\cite{DBLP:journals/corr/abs-1904-04346}, which first normalized the video into 96 frames, 
and then divided them into 6 segments where each clip contained 16 frames\footnote{We used I3D model~\cite{DBLP:conf/cvpr/CarreiraZ17} as the backbone. It took 16 frames as inputs.}.
%
%
In fact, as pointed in the recent work~\cite{DBLP:conf/iccv/JiaHuiaction}, 10 is more proper for the number of segments.
Based on this, we further studied two schemes. 
The first used $[0, 10, 20, 30, 40, 50, 60, 70, 80, 87]$ as the indices of beginning frames for the ten segments (denoted as ``10-seg-s1'').
Since $103/10 = 10.3$, we set the stride to be 10 in most cases. 
And the last beginning index was set to be 87 due to the length of the video is 103 frames.
The second scheme utilized $[0, 9, 19, 29, 38, 48, 58, 67, 77, 87]$ as the indices of beginning frames (denoted as ``10-seg-s2'').
As shown from the results, ``10-seg-s1'' achieves best result among the three.
Hence, we applied this scheme to the other actions in AQA-7 and MTL-AQA datasets in our paper.


\section{Visualization of Temporal Evolution}
In Section 4.3, we present a visualization result on the \textit{Gym}$\_$\textit{Vault} action~\cite{DBLP:conf/wacv/ParmarM19}. 
Here we further display two visualization results on the \textit{Diving} action~\cite{DBLP:conf/wacv/ParmarM19} in Figure \ref{fig:temporal}.
%
As it illustrates, the stage that the player enters the water plays a prominent part for action quality assessment.
For example, in the bottom instance, the player causes a large splash from the 8th segment to the 10th segment. 
Hence, for these segments, the distributions reach the peak at the low-level scores.

\end{appendices}

{\small
\bibliographystyle{ieee_fullname}
\bibliography{egbib}
}

\end{document}